\documentclass[11pt]{article}

\usepackage{graphicx}
\usepackage{booktabs}
\usepackage{algorithm}
\usepackage{algpseudocode}
\usepackage[T1]{fontenc}
\usepackage{lmodern}
\usepackage[flushleft]{threeparttable}
\usepackage[hidelinks]{hyperref}
\hypersetup{hidelinks}
\usepackage{multirow}
\usepackage{longtable}
\usepackage[margin=3cm]{geometry}
\usepackage{amsmath}
\usepackage{authblk} 

\title{Combining human parsing with analytical feature extraction and ranking schemes for high-generalization person reidentification}
\author[1]{Nikita Gabdullin}
\affil[1]{Joint Stock "Research and production company "Kryptonite"
\\E-mail: n.gabdullin@kryptonite.ru}
\date{}
\begin{document}

    \maketitle

    \begin{abstract}
        Person reidentification (re-ID) has been receiving increasing
        attention in recent years due to its importance for both science and
        society. Machine learning and particularly Deep Learning (DL) has become
        the main re-id tool that allowed researches to achieve unprecedented
        accuracy levels on benchmark datasets. However, there is a known problem
        of poor generalization of DL models. That is, models trained to achieve
        high accuracy on one dataset perform poorly on other ones and require
        re-training. To address this issue, we present a model without trainable
        parameters which shows great potential for high generalization. It
        combines a fully analytical feature extraction and similarity ranking
        scheme with DL-based human parsing used to obtain the initial subregion
        classification. We show that such combination to a high extent eliminates
        the drawbacks of existing analytical methods. We use interpretable color and texture features which have
        human-readable similarity measures associated with them. To verify the
        proposed method we conduct experiments on Market1501 and CUHK03 datasets
        achieving competitive rank-1 accuracy comparable with that of DL-models.
        Most importantly we show that our method achieves 63.9\% and 93.5\%
        rank-1 cross-domain accuracy when applied to transfer learning tasks. It is significantly higher than previously reported
        30-50\% transfer accuracy. We discuss the potential ways of adding new
        features to further improve the model. We also show the advantage of
        interpretable features for constructing human-generated queries from verbal
        description to conduct search without a query image.
    \end{abstract}

    \emph{Keywords}: person reidentification, re-id, human parsing, analytical
    features, similarity ranking, generalization.

\section{Introduction}\label{introduction}

Person re-identification (re-ID) is becoming one of the most significant
research topics in computer vision and computational intelligence due to
its two-fold importance for both science and society. It focuses on
person identification across camera systems addressing the increasing
demand for public safety. The problem of person identification commonly
goes along with person detection and tracking tasks {[}1{]}. The
reidentification task is commonly formulated as follows: a person in a
query image is to be matched with a person in an image or images
obtained from data streams of different cameras or a given camera at
various moments in time. A database of images or videos is often used
instead of live streams. The main challenge is to assess the similarity
between objects while taking into account possible changes in human appearance due to variations in
camera viewpoints, lighting conditions, person's pose, and occlusions.
In this paper we focus on the similarity assessment while for database
search techniques the reader is referred to literature {[}2{]} with
special mention of emerging graph-based methods {[}3{]}, {[}4{]}.

In practice re-ID person similarity evaluation is done using images
obtained from video frames. Recently various video re-ID approaches
emerged that effectively used temporal information to improve the
assessment accuracy {[}5{]}, {[}6{]}. Nevertheless, visual similarity
estimation remains the core of re-ID and methods that perform well on
images can be expected to perform even better when combined with
temporal data.

Fast development in the area of convolutional neural networks (CNN) and deep learning
(DL) that affected numerous areas of computer science also led to the
emergence of machine learning and DL-assisted re-ID techniques {[}7{]},
{[}8{]}. Re-ID requires CNNs to assess similarity for image pairs which
is not typical for conventional image classification tasks. This
inspired researchers to propose novel network architectures such as
Siamese networks {[}9{]}, {[}10{]}, specialized loss functions e.g.
triplet loss {[}11{]}, {[}12{]}, specific attention modules {[}13{]},
re-ID graph neural networks {[}14{]}, and others. Augmented images were
often included in training datasets to simulate illumination-related
effects {[}15{]}, {[}16{]}. Many early works focused on using CNNs for
either feature extraction or metric learning {[}17{]}, {[}18{]}, with
end-to-end models gradually becoming dominant in the field.

However, generalization of the results is challenging for DL-assisted
re-ID. That is, models trained on specific datasets tend to perform
poorly on other data, as can be illustrated by models trained on
Market1501 to 98\% rank-1 accuracy reaching only 38\% accuracy on
DukeMTMC and 51\% vice versa {[}19{]}. The effects of negative transfer
learning can be drastic due to significant variations in data contents,
e.g., different clothes people wear, different camera types, different
filming environments. A combination of such effects is often referred to
as ``Open-World re-ID problem'' addressing which requires further
complication of the CNN architecture {[}18{]}, {[}20{]}. Therefore,
models become extremely bulky with significant parameter redundancy that
reduces speed and increases computational power and storage demands. Thus,
generalization implies overparameterization that increases costs and
does not guarantee good performance.

To address this issue, we focus on analytical techniques to obtain a
compact and fast-performing model to reduce computational costs and
improve generalization. This is achieved by constructing compact
interpretable object descriptors (feature vectors) combined with a
similarity ranking scheme. Analytical models, while falling out of favor
recently due to growing interest towards DL models, previously showed
promising results on re-ID tasks {[}21{]} {[}22{]}. This was mainly
achieved by constructing an ensemble of features while having a
trainable model that finds feature weights in a manner similar to metric
learning {[}21{]}, {[}23{]}. This approach is different to CNN
techniques that work on images with nearly no preprocessing and generate
feature vectors incomprehensible for human operators. On the contrary,
analytical feature extraction makes it possible to construct
human-readable features that improve interpretability of the results. In
this work we focus on color and texture features. When comparing two
objects, vector elements will represent the percentage of similarity with
respect to a specific feature.

A drawback of old analytical models was that feature vectors were
global, i.e., generated for the whole image possibly including
background and other objects. Thus, it was impossible to obtain features
of specific elements like clothes or hair color. Now such separation
into different elements can be achieved using human parsing
that allows one to divide an object into class-specific subregions. This
makes it possible to generate feature vectors for specific classes and
to assess their similarity class by class. The overall similarity is
combined of class similarities weighted by class importance. It acts as
a score to obtain a similarity ranking for query images with test images
and evaluate rank-r matching rate {[}7{]}, {[}24{]}, {[}25{]}.
Furthermore, we propose a model without trainable parameters that is
immune to dataset-dependent negative effects which emerge due to data
fitting techniques used by CNNs or trainable analytical models.

The rest of the paper is organized as follows: Section~\ref{methodology} discusses the
methodology, Section~\ref{experiments} summarizes experiments and results, Section~\ref{discussions}
presents the discussions, and Section~\ref{conclusions} concludes the paper.

\section{Methodology}\label{methodology}

\subsection{Overview of the method}\label{overview-of-the-method}

\begin{figure} 
  \centering
  \includegraphics[scale=0.32]{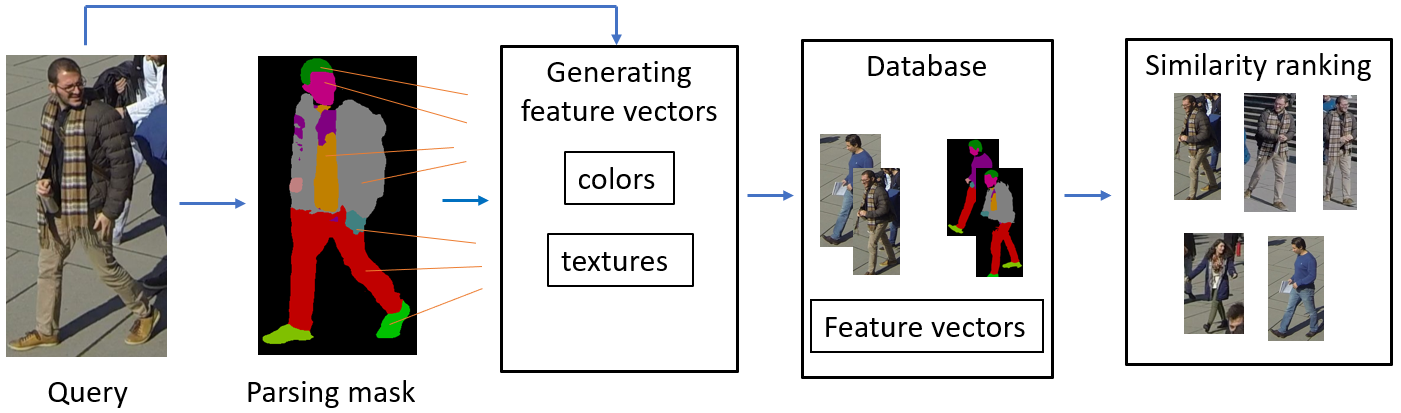}
  \caption{Method overview: parsing of the query image for per-class
  feature extraction and comparison with dataset images to form a
  similarity ranking.}
  \label{fig:1}
\end{figure}

The proposed method consists of four major steps shown in Fig.~\ref{fig:1}:
parsing of the query image, feature extraction for each subregion
(class), similarity score calculation between query and test set images
to calculate rank-r ranking results. It should be noted that for best
performance it is necessary to store parsing masks and feature vectors in
the database along with the original images. This allows image parsing
and feature extraction operations to be conducted once per image.

\subsection{Image parsing}\label{image-parsing}

Original image processing techniques utilized in analytical re-ID were
applied to full images, primarily due to the lack of reliable
segmentation techniques. This resulted in feature vectors of different
subregions being mixed complicating the similarity estimation.
Implementation of preliminary image parsing, and specifically human
parsing, is a promising technique that allows to divide an object into
subregions corresponding to specific classes. Whereas clustering-based
segmentation was well-known, it was computationally expensive and did
not provide sufficient accuracy. Spatiotemporal features based on graph
partitioning of the full image were proposed to localize feature vectors
to specific regions {[}22{]}. More recently a simple approach was
proposed that leveraged the fact that body parts in different images
ordinarily appeared in the same image sections in re-ID datasets. This
allowed one to separate images into three or more subregions and to
calculate local feature vectors for horizontal strips
{[}26{]}--{[}28{]}. This way the information about the head, torso and
legs could be obtained. Whereas being computationally convenient, this
method was not robust in scenarios that included pose change and partial
occlusion. Furthermore, the information regarding the difference between
elements withing one strip, like hands and torso, was lost. This was
addressed by several modifications based on more detailed representation
of human body structure {[}5{]}, {[}29{]}. Nevertheless, such naïve
segmentation has fundamental disadvantages which human parsing avoids.

The main goal of human parsing is to provide a mask which assigns a
specific class to every image pixel. For re-ID purposes these include
body parts, hair, clothes, and other wearables. Whereas this task can
also be performed by more sophisticated general purpose image
segmentation tools, they are unnecessarily complex due to redundant
classes such as furniture, animals, or vehicles which are irrelevant for
person segmentation {[}30{]}. Human parsing is usually performed by
neural networks trained on human parsing datasets, e.g. LIP or Pascal
{[}31{]}, {[}32{]}. Several researches have shown that incorporating
human parsing into re-ID framework significantly improves the prediction
accuracy {[}25{]}, {[}33{]}. However, existing solutions still utilize
additional neural networks for feature extraction, metric learning,
feature importance estimation (attention), and decision making resulting
in complex architectures.

In this paper we propose to combine recent advances in human parsing
with analytical feature extraction. Human parsing naturally handles
aspects that are challenging for analytical methods by providing the
initial segmentation of an object into subregions. Since every subregion
is assigned its class, this allows to compare feature vectors of the
same class without dealing with the noise originating from neighboring
subregions. Shape of the subregions mimics the real shape of the
elements in the image which pairs well with methods that do not require
to apply filters to subregions for feature extraction.

In this work we use out-of-the-box human parser SCHP {[}34{]} trained on
LIP dataset to obtain a parsing mask and generate color and texture
feature vectors for every class. It should be stressed that the parser
is used ``as is'' and no additional training has been performed. For a
pair of images feature vectors are compared only for classes present in
both images. It should be noted that whereas mean intersection over
union (mIOU) accuracy of SCHP is 59\%, it performed very well in our
experiments providing sufficiently accurate parsing masks. This is partially because
the accuracy requirements we pose on parsing for re-ID are different
from the ones in pure human parsing tasks. There is a common problem
that semantically similar classes, e.g. coats and upper clothes, are
easily mixed up. Whereas knowing whether a person is wearing a coat, or
a shirt, is indeed valuable and additional logic can be built using this
knowledge, it currently cannot be relied upon, and we do not require the
parsers to make such distinctions correctly. To address this, we merge
several semantically similar classes. Specifically, using LIP class
notation, we merge upper clothes, dress, coat, and jumpsuit into ``upper
clothes'' (class 5); and pants and skirt into ``pants'' (class 9). We also disregard the ``background'' class.  Thus,
we work with fifteen unique classes out of twenty original LIP ones.

\subsection{\texorpdfstring{Color similarity
}{2.3. Color similarity }}\label{color-similarity}

\subsubsection{\texorpdfstring{Choice of color space and histogram
modification}{Choice of color space and histogram modification}}\label{choice-of-color-space}

Whereas RGB color scheme, or color space, is widely used in image
encoding, it is not particularly suitable for image processing for
identification purposes. Therefore, other color spaces are used instead which
commonly include HSV, CMYK, YCbCr, and others {[}21{]}, {[}35{]}. HSV
color space and specifically \emph{H} channel information was found to
be the most descriptive by several researchers working on color
similarity for re-ID {[}15{]}, {[}35{]}. However, highest accuracy
results were commonly obtained using an ensemble of features, which
could include a variety of different channels from different color
spaces {[}21{]}. 

There is also a problem that these color spaces are perceptually
non-uniform with non-metric distances. On the contrary, CIE-Lab (Lab)
color space is a uniform color space where Euclidian distance can be used as a metric
for color difference calculation {[}36{]}. Whereas less popular in comparison to other color
spaces, it previously was successfully applied to image similarity estimation
{[}37{]}. In this paper we use Lab color space to leverage the
combination of perceptual uniformity with metric distance measurement to
create a two-fold color similarity estimate by comparing histograms of
Lab channels. It is also essential that the effects of illumination
changes which complicate color comparison are localized in the lightness
(\emph{L}) channel. We propose two approaches to illumination change
handling based on \emph{L} channel histogram analysis. They were
developed while working with Wildtrack dataset {[}38{]}.

\begin{figure} 
  \centering
  \includegraphics[scale=0.29]{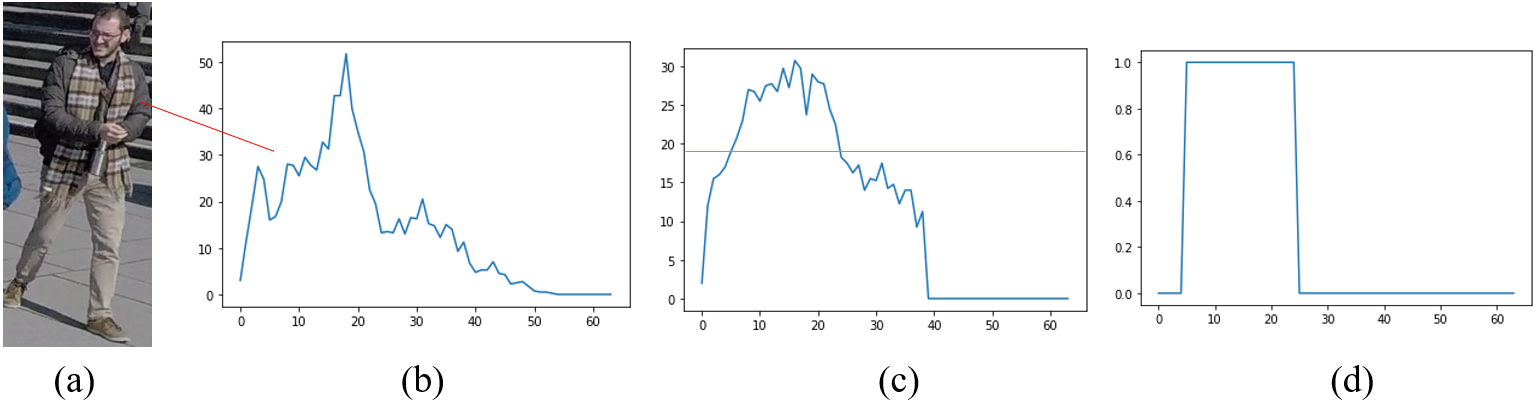}
  \caption{Histogram modification and binarization: (a) query image; (b) \emph{L} channel histogram
  of one of the classes (e.g., upper clothes); (c) modified histogram;
  (d) binarized histogram after thresholding.}
  \label{fig:2}
\end{figure}

Firstly, \emph{L} channel histograms (Fig.~\ref{fig:2} (b)) extracted for every
subregion class of an input image (Fig.~\ref{fig:2} (a)) are ``stretched''. To do
that the number of bins in the original histogram is first reduced from
256 to 64 by averaging the pixel values of neighboring bins. Then
Algorithm 1 is used to obtain a histogram shown in Fig.~\ref{fig:2} (c).

\begin{algorithm}
\caption{\emph{L} channel histogram modification}
\label{alg:1}
\begin{algorithmic}[1]
\Require \emph{h} is a 64 bin \emph{L} channel histogram 
\State Calculate \emph{h\textsubscript{av}} average of \emph{h} 
\State Zero bin \emph{i} if \emph{h\textsubscript{i}} < \emph{h\textsubscript{av}} for all bins
\State Calculate new average \emph{h\textsuperscript{'}\textsubscript{av}}
\State $\emph{m} \gets argmax(\emph{h})$ 
\For{\texttt{\emph{i} > \emph{m}}}
  \If{\emph{h}(\emph{i}) > \emph{h\textsuperscript{'}\textsubscript{av}}}
    \State $E \gets 0.5(\emph{h}(\emph{i}) - \emph{h\textsuperscript{'}\textsubscript{av}})$
    \State $\emph{h}(\emph{i}) \gets \emph{h}(\emph{i}) - E$
    \State add 0.25E each to the next four bins
  \EndIf
\EndFor
\State Repeat for \texttt{\emph{i} < \emph{m}}
\State Zero the first and the last non-zero bins
\end{algorithmic}
\end{algorithm}

This allows to normalize lightness levels among images taken at
different lighting conditions. Algorithm 1 is different from
conventional histogram stretching in that a) it does not force the new
histogram to occupy the whole range and b) the range of ``stretching''
depends on the extent that the number of pixels in certain bins exceeds
the average pixel number. In other words, the range of stretching is
proportional to the excess of pixels in some bins.

Before stretching a check for over-highlights is also performed. It was
noticed that for images where the number of pixels in the last bin
exceeds 1\% of all pixels prior to the proposed modification, \emph{L}
channel histograms become unreliable for color comparison due to loss of
information. Over-highlighted classes are not included into similarity
estimation.

\begin{figure} 
  \centering
  \includegraphics[scale=0.40]{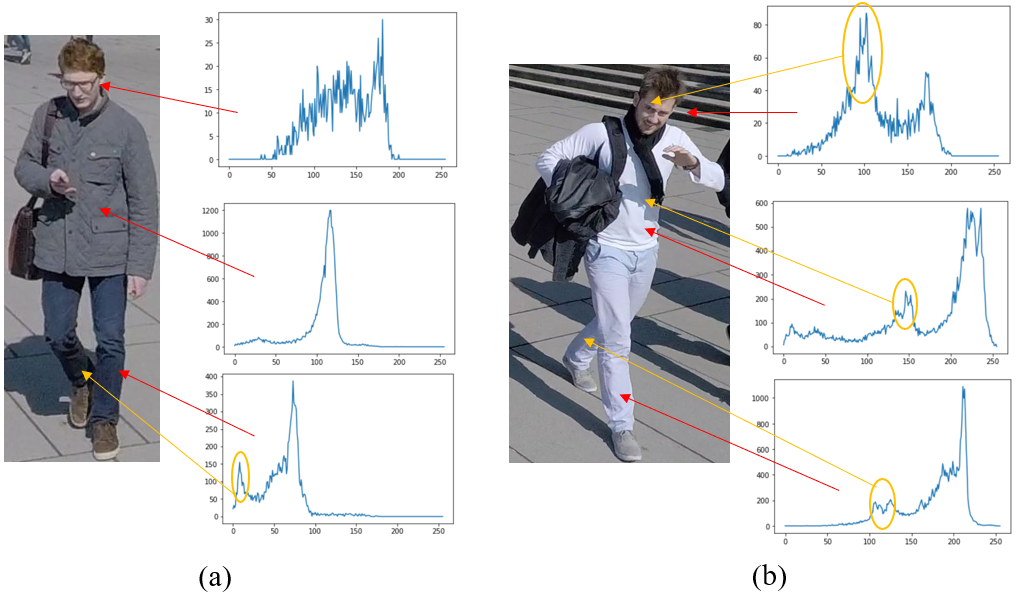}
  \caption{The correlation between shadows and peaks in \emph{L}
  channel histograms of face, upper clothes, and pants subregions: (a)
  prominent shadows appear only on pants; (b) shadows appear in every
  subregion.}
  \label{fig:3}
\end{figure}

Secondly, shadows can lead to significant distortion of color similarity
estimation. Local shadows appearing due to local occlusions further
complicate the situation since such effects cannot be handled by the
first approach. It was noticed that in many cases local shadows form
a distinct peak on \emph{L} channel histogram which precedes the peak
related to the most prominent \emph{real} color, as shown in Fig.~\ref{fig:3}.
This suggests that removing the first peak might significantly improve
the analysis accuracy. However, regions related to shadows are harder to
locate for low-resolution or poorly lit images. Naively removing a part
of the histogram results in worse accuracy in such cases. The search for
a comprehensive approach to handling local shadows is still being
conducted.

\subsubsection{Representative color intensities and histogram
thresholding}\label{representative-color-intensities-and-histogram-thresholding}

Possible changes in view angle present another challenge for color
comparison methods. They can lead to same distinct areas appearing to
have different sizes, making Lab channels' histograms for same classes
of the same object shift dramatically. This makes bin-to-bin histogram
comparison methods less accurate which has inspired the development of
other comparison metrics {[}37{]}, {[}39{]}. However, they still might
give misleading results due to drifts in pixel intensities caused by
illumination and point of view changes. To address this issue, we
propose a simple thresholding scheme with a purpose of finding the most
representative pixel intensities.

Firstly, the number of bins in color channel histograms is reduced from
256 to 64 as for \emph{L} channel histograms in Section~\ref{choice-of-color-space}. This
reduces the influence of drifting effects. Secondly, for every bin
\emph{j} of the original histogram \emph{h\textsubscript{o}} of channel
\emph{i} (where \emph{i} = \emph{L,a,b)}, where \emph{L} channel
histogram is ``stretched'', we assign ``1'' in its \emph{binarized}
version \emph{h\textsubscript{b}} when the number of pixels in that bin
exceeds the threshold \emph{k\textsubscript{bi}}, and assign ``0''
otherwise as shown in Fig.~\ref{fig:2} (d) so that

\begin{equation}
  h_{bij} = \left\{ \begin{matrix}
  1\ if\ h_{oij} \geq k_{bi}, \\
  0\ otherwise. \\
  \end{matrix} \right.\ 
  \label{eq:hbij}
\end{equation}

\begin{equation}
  k_{bi} = k_{inc} \frac{\sum_{j = 0}^{255}h_{bij}}{256},
  \label{eq:kbi}
\end{equation}

where \emph{k\textsubscript{inc}} is an empirical coefficient that
increases threshold value over simple average. In our experiments
\emph{k\textsubscript{inc}} = 1.5 was found to provide the best results.
This approach also allows us to design a comparison scheme for a pair of
histograms. Given two binarized histograms \emph{h\textsubscript{bi1}}
and \emph{h\textsubscript{bi2}} of the same class in two images we find
their sum \emph{h\textsubscript{ti}} which can consist only of zeroes,
ones, and twos and define color channel similarity measures as:

\begin{equation}
  h_{ti} = h_{bi1} + h_{bi2},
  \label{eq:hti}
\end{equation}

\begin{equation}
  S_{i} = \frac{s_{2i}(h_{ti})}{s_{1i}(h_{ti}) + s_{2i}(h_{ti})},
  \label{eq:Si}
\end{equation}

where \emph{s\textsubscript{1}} and \emph{s\textsubscript{2}} are the
numbers of times ``1'' or ``2'' appeared in \emph{h\textsubscript{t}},
respectively. The intuition behind this is to find intensities prominent
in both histograms as well as their fraction relative to the total
number of non-zero bins so that \emph{S\textsubscript{i}} lies in
{[}0,1{]} range. This operation is very similar to histogram
intersection {[}39{]}. However, it does not require
\emph{h\textsubscript{bi1}} and \emph{h\textsubscript{bi2}} to have the
same number of non-zero bins and it is commutation-invariant. The former
is very important since the number of non-zero intensities after
thresholding can vary significantly for different images.

\subsubsection{Distance in Lab color space as similarity
measure}\label{distance-in-lab-color-space-as-similarity-measure}

It was previously mentioned that distances in Lab color space obey
Euclidian distance equation. This allows to judge how similar two colors
are depending on how close their corresponding Lab vectors lie. Since we
deal with histograms rather than actual colors, we first propose to
calculate average intensity for histograms in every channel, and to
treat such triplet as coordinates of a point in Lab space. In this case
numbers of pixels \emph{h\textsubscript{oij}} in bins \emph{j} of the
original histogram \emph{h\textsubscript{o}} act as weights so average
intensity in color channel \emph{i} is calculated as

\begin{equation}
  I_{i} = \frac{\sum_{j = 0}^{255}{j h}_{oij}}{\sum_{j = 0}^{255}h_{oij}}.
  \label{eq:Ii}
\end{equation}

For image segments of uniform color a triplet (\emph{I\textsubscript{L},
I\textsubscript{a}, I\textsubscript{b}}) is expected to lie close to the
main color perceived by human eye. In case of multiple subregions with
different colors it may not correspond to any real color present in the
image. However, this is sufficient for distance calculation purposes, so
for two segments of the same class

\begin{equation}
  d = \ \sqrt{\left( I_{L1} - I_{L2} \right)^{2} + \left( I_{a1} - 
  I_{a2} \right)^{2} + \left( I_{b1} - I_{b2} \right)^{2}}.
  \label{eq:d}
\end{equation}

We then propose a normalized similarity measure
\emph{S\textsubscript{d}} with values in {[}0,1{]} range which is
related to \emph{d} as

\begin{equation}
  S_{d} = \left\{ \begin{matrix}
  1 - \frac{d}{k_{d}}\ if\ d < k_{d}, \\
  0\ otherwise, \\
  \end{matrix} \right.\ 
  \label{eq:Sd}
\end{equation}

where \emph{k\textsubscript{d}} is an empirical distance coefficient and
in this work \emph{k\textsubscript{d }}= 35.

\subsection{Texture similarity}\label{texture-similarity}

Analytical texture descriptors are often separated into two broad
categories: skeletal shape abstraction and local texture descriptors.
The former deals primarily with the overall geometrical structure and
object's shape {[}40{]}. The latter extracts subregion-specific
information about the object's texture using filters or operations on
pixels and their neighborhoods {[}41{]}.

In this work we focus on the local texture description because image
parsing allows us to obtain initial classification of subregions. In our
case skeletal descriptors provide insufficient information since two
objects that are already classified as ``upper clothes'' or ``pants'' by
default have very similar structure. In prior work filter packs that included Gabor filters
were found to perform well on re-ID tasks {[}21{]}. Whereas such filters
provided desirable translation and affine-invariant texture
descriptors, the necessity to fine-tune filter parameters led to a
significant number of filters required to analyze a sufficiently vast
variety of different textures. To avoid this, we have chosen Local
Binary Pattern (LBP) method due to its ability to generate texture descriptors.

LBP is a local texture descriptor that was found to be extremely useful
in texture and face classification. It assigns every pixel of a
grayscale image a binary number with bits representing whether pixel
neighbor's intensity is greater than that of a central pixel. In the
simplest case of only checking the closest neighbors (radius \emph{r}=1)
we obtain an 8-bit binary number and convert it to decimal. All decimal
numbers of the image are combined into a 256-bin histogram. It is
important to note that every bin corresponds to a unique texture
\emph{type}, which can be significantly different for neighboring bins.

Over the years various modifications to the original LBP were proposed,
including multiscale extension {[}42{]}, spatial enhancement {[}43{]}, and 
texture uniformity classification {[}44{]}. These methods often include
concatenation of several local histograms to encode additional spatial
information. However, such modifications are not suitable for re-ID
purposes since they are not translation and rotation invariant. This is
acceptable for face recognition tasks where distinct facial features are
always localized in the same regions of an image, but not for re-ID
images, where point of view and object orientation can change
dramatically. The separation of textures into \emph{uniform} and
\emph{non-uniform} classes that was found useful in many face
recognition tasks also did not perform well here. The reason lies in the
significant reduction in the number of non-zero bins and loss of some
descriptive information. Whereas for face recognition tasks uniform
patterns were found most informative, this appears not to be the case
for re-ID tasks.

Therefore, we use the original LBP method with one slight modification.
It was shown that LBP can be made rotation and affine transformation
invariant by performing a circular bit-wise right shift on the binary
numbers {[}45{]}. Whereas this also reduces the number of possible
non-zero bins, these invariances are essential for re-ID. This
modification was found to improve the performance in our experiments.

In this work we calculate two LBP histograms for every class of interest
in the image. They correspond to LBP descriptions of subregion's contour
and its inner area. It was found that separating them positively affects
the similarity estimation because contour similarity varies in a wider
range and its variations are not trivial with respect to changes in the
inner area's similarity. As a similarity measure for contour or inner
area (channel \emph{m = co, in}) histograms of two images we use
histogram intersection. In order to obtain a commutation invariant
similarity measure for a pair of histograms, the histograms are first
normalized with respect to the total number of points in all bins in
that histogram yielding modified histograms \emph{h\textsubscript{tm1}}
and \emph{h\textsubscript{tm2}}, so that values in bins show percentages
of all textures corresponding to those bins, so that

\begin{equation}
  S_{m} = \ \sum_{j = 0}^{255}{min(h_{tm1j},h_{tm2j})}
  \label{eq:Sm}
\end{equation}

with values in {[}0,1{]} range. In our experiments similarity estimates
obtained using texture and color descriptors did not provide overlapping
results implying that there is no trivial relationship between features.
This is essential because texture descriptors can address cases
challenging for color similarity estimation and vice versa.

\begin{table}
  \caption{Feature weights \emph{w\textsubscript{f}} that determine
  feature importance in class similarity evaluation.}
  \label{tab:1}
  \bigskip
  \centering
  \begin{tabular}{|c|c|c|c|c|c|c|}
    \hline
    Feature & \emph{L} & \emph{a} & \emph{b} & \emph{d} &
    \emph{t\textsubscript{in}} & \emph{t\textsubscript{co}} \\
    \hline
    Weight & 0.13 & 0.13 & 0.13 & 0.31 & 0.15 & 0.15 \\
    \hline
  \end{tabular}
\end{table}

\begin{table}
  \caption{Class \emph{w\textsubscript{c}} weights determining class
  importance in similarity score calculation.}
  \label{tab:2}
  \bigskip
  \centering
  \begin{tabular}{|c|c|c|c|c|c|}
    \hline
    Parsing class & Hair, socks, face, & Hat, gloves, sun- & Scarf & Pants &
    Upper clothes  \\
    & legs, arms & glasses, shoes &  &  & \\
    \hline
    Weight & 1 & 2 & 3 & 6 & 8 \\
    \hline
  \end{tabular}
\end{table}

\subsection{Similarity score calculation for similarity
ranking}\label{similarity-score-calculation-for-similarity-ranking}

Previous sub-sections discuss six feature channels
(\emph{S\textsubscript{f}}) with similarity measures
\emph{S\textsubscript{L}, S\textsubscript{a}, S\textsubscript{b},
S\textsubscript{d}, S\textsubscript{co}}, and
\emph{S\textsubscript{in}}. In general, for a class \emph{c} with
multiple feature channels class similarity is calculated as

\begin{equation}
  S_{c} = \ \sum_{f}^{}{w_{f} S_{f}}
  \label{eq:Sc}
\end{equation}

\begin{equation}
  \sum_{f}^{}w_{f} = \ 1
  \label{eq:sumwis1}
\end{equation}

where \emph{w\textsubscript{f}} is weight of a feature channel \emph{f}.
Table~\ref{tab:1} shows \emph{w\textsubscript{f }}values used in this study. It is
worth pointing out that their sum equals one which ensures that
\emph{S\textsubscript{c}} is in {[}0,1{]} range. For a query-test pair
with \emph{n} shared classes the total similarity score is calculated as

\begin{equation}
  S_{sim} = \ \sum_{c}^{n}{w_{c} S_{c}}
  \label{eq:Ssim}
\end{equation}

\begin{equation}
  S_{simn} = \ \frac{S_{sim}}{\sum_{c}^{n}w_{c}}
  \label{eq:Ssimn}
\end{equation}

where \emph{w\textsubscript{c}} are class weights shown in Table 2.

It should be noted that unlike channel-wise similarity estimates,
similarity score \emph{S\textsubscript{sim}} does not naturally fall in
{[}0,1{]} range. In general, \emph{S\textsubscript{sim}} depends on the
number of shared classes \emph{n}, which varies from one to the number
of classes in query image. The highest score would be obtained by a pair
of images with all possible classes having perfect similarity among all
features, which for the weights in Table~\ref{tab:2} amounts to 34. In practice it
is extremely rare to have objects with all fifteen different classes present
in both images and similarity scores are much lower. Whereas a
percentage similarity measure can be obtained using \eqref{eq:Ssimn}, it is less
meaningful as similarity score since an image pair with fewer classes
with low \emph{w\textsubscript{c}} but high \emph{S\textsubscript{c}}
will have higher \emph{S\textsubscript{simn}} than another pair with many
important classes but slightly lower \emph{S\textsubscript{c}}. Thus, we
propose to use \emph{S\textsubscript{sim}} rather than
\emph{S\textsubscript{simn}} as scores for ranking.

To calculate rank-r matching rate for a query image its scores
\emph{S\textsubscript{sim}} with all test images are calculated and sorted
in descending order with \emph{r} highest score test images returned for
match checking

\begin{equation}
  rank\ r = \ \left\{ \begin{matrix}
  1\ if\ true\ match\ in\ r \\
  0\ otherwise \\
  \end{matrix} \right.\
  \label{eq:topr}
\end{equation}

Most popular rank-r metrics are rank-1, rank-5, and rank-10. To evaluate
method performance on a dataset with multiple queries the rank-r
accuracy is averaged for all queries. The main disadvantage of rank-r
metric is that it requires a single match in \emph{r} which is usually
the simplest among possible tests. Thus, it does not contain information
about other possible matches and does not reflect the ability of the
method to capture complicated matches. To address this problem, mean
average precision (mAP) metric is often used along with rank-r when
there is more than one correct match in the test set {[}7{]}, {[}25{]}.
It is calculated as

\begin{equation}
  mAP = \ \frac{1}{n_{q}}\sum_{i}^{}{\frac{1}{n_{TPi}}\sum_{j}^{}\frac{m_{ij}}{j}}
  \label{eq:mAP}
\end{equation}

\begin{equation}
  m_{ij} = \ \left\{ \begin{matrix}
  cumulative\ number\ of\ true\ matches\ for\ query\ i\ found\ at\ step\ j \\
  0\ otherwise \\
  \end{matrix} \right.\
  \label{eq:mij}
\end{equation}

where \emph{n\textsubscript{q}} is number of queries in query set,
\emph{n\textsubscript{TPi}} is number of true matches for
\emph{i\textsuperscript{th}} query in test set.

\section{Experiments}\label{experiments}

\subsection{\texorpdfstring{Details of the datasets
}{3.1. Details of the datasets }}\label{details-of-the-datasets}

To verify our methods we conduct experiments on two datasets: Market1501
{[}28{]} and CUHK03 {[}46{]}. For the latter we specifically use a
version with a clear query-test separation {[}47{]}. We conduct
experiments on both labeled (L) and detected (D) versions of CUHK03.
Table~\ref{tab:3} provides an overview of the datasets.

Whereas Table~\ref{tab:3} shows that Market1501 is significantly larger than
CUHK03, the latter provides a considerably bigger challenge due to poor
lighting conditions which make different persons look similar. CUHK03 is
considered to be one of the most challenging re-ID dataset by some
researchers {[}48{]}. This explains lower performance of our and other
methods on CUHK03 compared to Market1501.

\begin{figure}[t]
  \centering
  \includegraphics[scale=0.45]{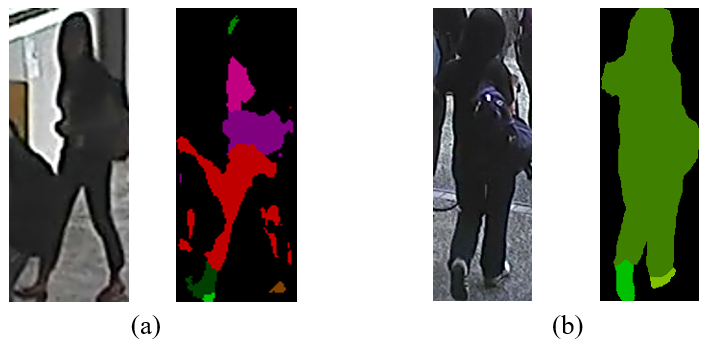}
  \caption{Parsing errors: (a) incorrect parsing and labeling of
  subregions, (b) overall correct parsing but most subregions are
  mislabeled as ``jumpsuit''.}
  \label{fig:4}
\end{figure}

We also inspected SCHP-generated parsing masks of all queries and marked
samples where parser made considerable mistakes. Figure \ref{fig:4} illustrates
typical parsing errors where a parsing mask is obviously incorrect, or a
significant number of subregions are mislabeled. For some experiments
summarized in Tables \ref{tab:4} and \ref{tab:5} we excluded query images with parsing
errors while labeling the remaining queries as clear queries (cq). We
provide the results for both full and clear queries to show the effect
of parsing errors on the overall accuracy, and to better investigate the
accuracy of the proposed feature extraction and similarity ranking
approaches. It should be noted that parsing errors have not been removed
from test subsets.

\begin{table}
  \caption{The details of the used datasets.}
  \label{tab:3}
  \bigskip
  \centering
  \begin{tabular}{|c|c|c|c|c|c|c|}
    \hline
    Dataset & Identities & Images & Test & Queries &
    Clear & Cameras  \\
    & & &  &  & queries & \\
    \hline
    Market1501 & 1501 & 32668 & 19732 & 3368 & 3062 & 6 \\
    \hline
    CUHK03 (L) & 1360 & 13164 & 5328 & 1400 & 1310 & 2 \\
    \hline
    CUHK03 (D) & 1360 & 12697 & 5332 & 1400 & 1294 & 2 \\
    \hline
  \end{tabular}
\end{table}

\subsection{Market1501 experiments}\label{market1501-experiments}

\begin{table}
  \caption{Performance comparison on Market1501 dataset.}
  \label{tab:4}
  \bigskip
  \centering
  \begin{tabular}{|c|c|c|c|c|c|c|}
    \hline
    Model & Backbone & Human & Learning & Rank-1 &
    Rank-10 & mAP  \\
    & & parsing & type & & & \\
    \hline
    SML {[}49{]} `19 & ResNet-50 & No & US* & 67.7 & - & 40 \\
    \hline
    SIV {[}50{]} `17 & ResNet-50 & No & S & 79.51 & - & 59.87 \\
    \hline
    MSCAN {[}51{]} `17 & Custom & No & S & 80.31 & - & 57.53 \\
    \hline
    CAP {[}52{]} `21 & ResNet-50 & No & US & 91.4 & 97.7 & 79.2 \\
    \hline
    SSP {[}53{]} `18 & ResNet-50 & Yes & S & 92.5 & - & 80 \\
    \hline
    SPReID {[}25{]} `18 & Inception & Yes & S & 94.63 & 98.4 & 90.96 \\
    \hline
    Pyramid {[}54{]} `19 & ResNet & No & S & 95.7 & 99 & 88.2 \\
    \hline
    APNet-C {[}19{]} `21 & ResNet-50 & No & S & 96.2 & - & 90.5 \\
    \hline
    CTL-S {[}55{]} `21 & ResNet-50 & No & S & 98 & 99.5 & 98.3 \\
    \hline
    Ours & ResNet-101 (parser) & Yes & A & 91 & 96 & 25.2 \\
    \hline
    Ours (cq) & ResNet-101 (parser) & Yes  & A  & 93.5 & 98.0 & 25.3 \\
    \hline
  \end{tabular}
  \small
  *S -- supervised, US -- unsupervised, A -- analytical, no learning
\end{table}

\begin{figure} 
  \centering
  \includegraphics[scale=0.45]{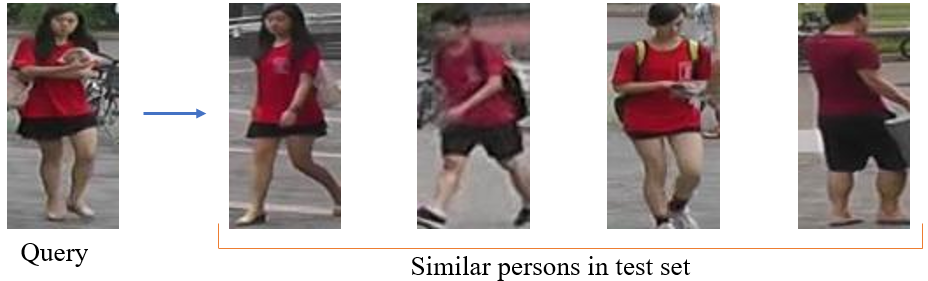}
  \caption{Similarly dressed people (right) in test set of Market1501
  dataset which have similar scores for a query (left) resulting in low
  mAP.}
  \label{fig:5}
\end{figure}

Table~\ref{tab:4} shows a comparison of accuracy between our method and other
models on Market1501 dataset. It shows that we achieve results
comparable with unsupervised learning models and DL parsing models in
rank-1 and rank-10 categories. However, mAP accuracy is significantly
lower in comparison with other approaches. Fig.~\ref{fig:5} shows an example of a
typical challenge wherein two different people who are dressed similar
have almost the same similarity scores, so their images are mixed 
obstructing the correct person retrieval. This behavior is
understandable since we only use color and texture as features with
clothes providing the greatest contribution. It should be noted that we
still outperform modern unsupervised models on rank-1/rank-10 metrics.

\subsection{CUHK03 experiments}\label{cuhk03-experiments}

Table~\ref{tab:5} shows trends similar to Table~\ref{tab:4}. However, the rank-1 accuracy
achieved on CUHK03 is lower than that of current cutting-edge models by
a bigger margin compared to Market1501 experiments. This can be
explained by poor lighting resulting in color becoming a less reliable
criterion. The effects shown in Fig.~\ref{fig:5} are also common for CUHK03
dataset. However, we still outperform several older DL models. It should
be stressed that the results in Table~\ref{tab:5} are obtained using exactly the
same model as the results in Table~\ref{tab:4}, and the importance of this fact is
discussed further in Section~\ref{generalization-and-potential}.

\begin{table}
  \caption{Performance comparison on CUHK03 dataset.}
  \label{tab:5}
  \bigskip
  \centering
  \begin{tabular}{|c|c|c|c|c|c|c|c|}
    \hline
    Model & Backbone & Human & Learning & Rank-1 & mAP &
    Rank-10 & mAP  \\
    & & parsing & type & (L) & (L) & (D) & (D) \\
    \hline
    HA-CNN {[}56{]} `18 & Inception & No & WS* & 44.4 & 41 & 41.7 & 38.6 \\
    \hline
    DaRe {[}57{]} `18 & DenseNet-201 & No & S & 56.4 & 52.2 & 54.3 & 50.1 \\
    \hline
    DaRe {[}57{]} `18 & DenseNet-201 & No & S + RR & 73.8 & 74.7 & 70.6 & 71.6 \\
    \hline
    SSP {[}53{]} `18 & ResNet-50 & Yes & S & 65.6 & 63.1 & 66.8 & 60.5 \\
    \hline
    OSNet {[}58{]} `19 & OSNet & No & S & - & - & 72.3 & 67.8 \\
    \hline
    Top-SB-Net {[}48{]} `20 & ResNet-50 & Yes & S & 79.4 & 75.4 & 77.3 & 73.2 \\
    \hline
    Top-SB-Net {[}48{]} `20 & ResNet-50 & Yes & S + RR & 88.5 & 86.7 & 86.9 & 85.7 \\
    \hline
    MPN {[}59{]} '21 & ResNet-50 & No & S & 85 & 81.1 & 83.4 & 79.1 \\
    \hline
    Deep Miner {[}60{]} `21 & RedNet-50 & No & S & 86.6 & 84.7 & 83.5 &
    81.4 \\
    \hline
    LightMBN {[}61{]} `21 & OSNet & No & S & 87.2 & 85.1 & 84.9 & 82.4 \\
    \hline
    Ours & Resnet-101 (parser) & Yes & A & 61.1 & 20.9 &
    59.7 & 20.2 \\
    \hline
    Ours (cq) & Resnet-101 (parser) & Yes & A & 63.9 & 22.1 & 62.2 & 21.4 \\
    \hline
  \end{tabular}
  \small
  *S -- supervised, WS, - Weakly Supervised, RR -- re-ranking, A -- analytical, no learning
\end{table}

\section{Discussions}\label{discussions}

\subsection{\texorpdfstring{Performance and space requirements
}{4.1. Performance and space requirements }}\label{performance-and-space-requirements}

In this paper we propose a system which feature extraction and image
comparison modules are fully analytical without trainable parameters.
Such system has little hardware requirements and does not require GPU.
Feature vectors are also compact requiring little storage space. Due to
simplicity of the operations \eqref{eq:Sd}-\eqref{eq:Ssimn} the analysis takes milliseconds
even on an average-grade machine and supports multi-threading on a
multi-core CPU. The most time-consuming operation is feature extraction
since it requires one ``walk'' over the input matrix (query image).
However, it is only done once for every image as feature vectors are
stored for future comparison. The results shown in Section~\ref{experiments} are
comparable with some cutting-edge DL models indicating viability of the
proposed approach. Section~\ref{generalization-and-potential} also discusses much higher generalization
potential when the results in Tables \ref{tab:4} and \ref{tab:5} are analyzed together.

One might doubt the above claim since our approach still requires a
parser. In this paper we use SCHP with ResNet101 backbone (43 million
parameters) which is larger than backbones of most models in Tables \ref{tab:4}
and \ref{tab:5}. However, using such a large parser is not a necessary
requirement. A more compact network could be used instead with little
loss of accuracy as discussed in {[}34{]}. It was shown that
MobileNet-based SCHP parser (4.2 million parameters) performs only 5\%
worse that ResNet101-based one. At the same time MobileNet backbone is
about ten times more compact. Furthermore, it has been discussed in
Section~\ref{image-parsing} that the requirements on parsing accuracy for the proposed
re-ID approach are less strict compared to pure human parsing tasks.
This implies that for our purposes even more compact backbones can be
used, e.g. OSNet (2.2 million parameters) {[}58{]}. Future work will
elaborate on parser size minimization and investigate the possibility of
realizing the proposed re-ID system as an edge computing application.

\hypertarget{adding-new-features}{%
\subsection{Adding new features}\label{adding-new-features}}

A major contribution of this paper is feature extraction and similarity
score calculation scheme which is based on class similarity
considerations. In this paper we propose two types of features, i.e.,
color and texture features discussed in Section~\ref{methodology}. Whereas this is
sufficient to achieve high rank-1 accuracy on studied datasets, low mAP
accuracy indicates that further improvements are desired. This can be
achieved by adding new features on feature vector generation step in Fig.~\ref{fig:1} to
address the problem illustrated by Fig.~\ref{fig:5}.

There are two rules that a new feature should comply with: a) its
associated similarity measure \emph{S\textsubscript{f}} should be in {[}0,1{]} range and
b) its weight \emph{w\textsubscript{f}} should be added in Table~\ref{tab:1} and all weights
should be adjusted to comply with \eqref{eq:sumwis1}. Class similarity scores are
still calculated using \eqref{eq:Sc}. Hence, new features can be added with
minimal interference with the rest of the model. Future work will
consider pattern and shape features to improve the accuracy.

\subsection{Human-readable vectors and human-generated queries}\label{human-readable-vectors-and-artificial-query-generation}

In Section~\ref{methodology} we have discussed that all similarity measures are in {[}0,1{]}
range and they can be interpreted by a human operator as a percentage
similarity according to a specific feature. Whereas this
interpretability is useful for analysis purposes, it can also be used to
construct feature vectors while not having an actual image. The
possibility to conduct search for an object without query images was
investigated in early days of re-ID {[}35{]}, and it can be further
improved using our proposed approach which includes human parsing.

Constructing feature vectors can be done as follows. Let's assume that
colors of clothes of a person are known. RGB values of colors can be
converted into Lab to instantly obtain triplets
(\emph{I\textsubscript{L}, I\textsubscript{a}, I\textsubscript{b}})
discussed in Section~\ref{color-similarity}. Using same triplets binarized histograms can
be generated by assigning several ``ones'' in a region surrounding the
main color's intensity. LBP texture features are less intuitive, so a
look-up table of vectors of typical textures can be used instead.

Now let's consider a situation in which query is a vague verbal
description. Comparing such query with a database would be impossible
for methods that rely on convolutional operations, i.e., all DL models,
since the image is missing. Generating vectors corresponding to a given
verbal query is also impossible since feature vectors generated by DL
models have no clear interpretation. This would require to
generate an artificial image which is a very complex task and would
require making assumptions of numerous unknown parameters {[}62{]}.

\begin{figure} 
  \centering
  \includegraphics[width=\textwidth]{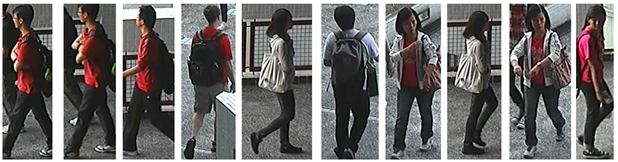}
  \caption{Search results for a human-generated query of a person wearing a
  red shirt, black pants, and black shoes in CUHK03 dataset.}
  \label{fig:6}
\end{figure}

\begin{figure} 
  \centering
  \includegraphics[width=\textwidth]{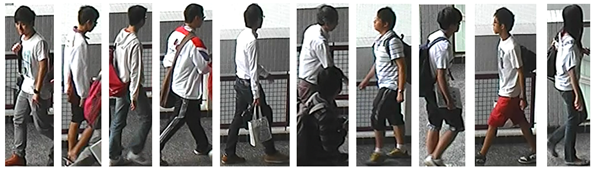}
  \caption{Search results for a request to find ten people wearing white
  shirts.}
  \label{fig:7}
\end{figure}

On the contrary, it is possible to search for matches for such query
using our proposed method. Figure \ref{fig:6} illustrates an example of a search
for a dark hair person wearing a red shirt, black pants, and black
shoes. For this search color descriptions are converted into vectors and
texture features are generated using values typical for clothes in CUHK03
dataset. Figure \ref{fig:7} illustrates another search results for a person
wearing a white shirt. Hence, an operator needs to describe only
relevant classes without the need to specify other class properties, e.g.,
shoes' color if it is unknown.

\subsection{Generalization and potential application to
Open-World
scenario}\label{generalization-and-potential}

Tables \ref{tab:4} and \ref{tab:5} illustrate that the highest accuracy is achieved by
models trained using supervised learning especially if re-ranking is
applied (see Table~\ref{tab:5}), which encourages models to give higher scores to
true matches, thus increasing mAP. This observation makes it tempting to
make parameters in Tables \ref{tab:1} and \ref{tab:2} trainable to possibly improve the
accuracy in our experiments. However, this would make generalization
worse due to the bias the model would develop towards training datasets.
Such effects are well-known in machine learning.

On the contrary, the absence of trainable parameters makes our approach
dataset-agnostic. Our results on Market1501 in Table~\ref{tab:4} are obtained
using the same model as in case of CUHK03 results in Table~\ref{tab:5}. Therefore,
this can be considered analogous to transfer learning experiments in
context of neural networks. Therefore, our 62.9\% Market-CUHK rank-1 ``transfer'' accuracy is significant when compared to
50.1\% DukeMTMC-Market transfer in {[}19{]}, as shown in Table~\ref{tab:6}. We
also achieve mAP which is comparable or even higher than the transfer
learning mAP of APNet-C. It should also be noted that DukeMTMC dataset
is less challenging than CUHK03 {[}59{]}, {[}60{]}, {[}63{]}. Therefore,
the proposed method is very promising from the perspective of the
development of highly generalizable models. This also suggests that when
accuracy improvements are achieved on one dataset it is likely to
improve the accuracy on other datasets, too. Such behavior, whereas
desirable, is not guaranteed for DL models. In contrast with the current
trends towards end-to-end DL models, this paper indicates significant
potential in combining machine learning (human parsing) with human
intelligence (analytical features) to obtain more flexible systems
that are more robust in Open-World scenarios while being easier for
human interpretation and understanding.

\begin{table}
\caption{Comparison between dataset and transfer accuracies between
supervised DL CNN models and the proposed approach.}
\label{tab:6}
\centering
\begin{tabular}{|c|c|c|c|c|c|c|c|c|c|c|}
  \hline
  Experiment & \multicolumn{2}{c|}{Market} &\multicolumn{2}{c|}{Duke} 
  &\multicolumn{2}{c|}{CUHK} & \multicolumn{2}{c|}{Market-X} & \multicolumn{2}{c|}{X-Market} \\
  \hline
  Metric & R1 & mAP & R1 & mAP & R1 & mAP & R1 & mAP & R1 & mAP \\
  \hline
  APNet-C {[}19{]} & \textbf{96.2} & \textbf{90.5} & 90.4 & 81.5 &
  \textbf{87.4} & \textbf{85.3} & 37.7 & \textbf{22.8} & 50.9 & 23.7 \\
  \hline
  Ours (cq) & 93.5 & 25.3 & - & - & 63.9 & 22.1 & \textbf{63.9} & 22.1 &
  \textbf{93.5} & \textbf{25.3} \\
  \hline
\end{tabular}
\begin{tablenotes}
  \small
  \item *\emph{X} is DukeMTMC for APNet-C, and CUHK03 in our experiments
\end{tablenotes} 
\end{table}

\section{Conclusions}\label{conclusions}

This paper proposes a re-ID system that combines analytical feature
extraction and similarity ranking scheme with human parsing. It shows
that parsing class masks provide sufficient information to overcome
known limitations of analytical re-ID methods. Two types of features,
namely Lab color and LBP texture features, are utilized and several
original pre-processing techniques are proposed. These features are
combined with class similarity evaluation scheme used to obtain a
similarity ranking comparable with conventional rank-r evaluation
metric. The obtained results show rank-1 accuracy comparable with
supervised DL models and exceed that of unsupervised ones on Market1501
and CUHK03 datasets. The reasons for reduced mAP accuracy are discussed
and potential solution in adding new feature channels is proposed. It is
shown that by having no trainable parameters the proposed model has
significant generalization potential as illustrated by 93.5\% and 63.9\%
rank-1 ``transfer'' accuracies which significantly exceed that of
conventional DL models. Future work will explore additional features to
improve accuracy along with parser size minimization to realize the
proposed system as an edge computing application.

\section*{Acknowledgement}\label{acknowledgement}

The author would like to thank his Kryptonite colleagues Dr Anton
Raskovalov and Dr Igor Netay for fruitful discussions and Vasilii
Dolmatov for his assistance in problem formulation, choice of
methodology, and supervision.

\section*{References}\label{references}

{[}1{]} L. Zheng, Y. Yang, and A. G. Hauptmann, ``Person
Re-identification: Past, Present and Future,'' vol. 14, no. 8, pp.
1--20, 2016, {[}Online{]}. Available: http://arxiv.org/abs/1610.02984.

{[}2{]} R. Iguernaissi, D. Merad, K. Aziz, and P. Drap, ``People
tracking in multi-camera systems: a review,'' \emph{Multimed. Tools
Appl.}, vol. 78, no. 8, pp. 10773--10793, 2019, doi:
10.1007/s11042-018-6638-5.

{[}3{]} E. Kodirov, T. Xiang, Z. Fu, and S. Gong, ``Person
Re-identification by Unsupervised Graph Learning,'' in \emph{Proc. ECCV},
2016, pp. 178--195.

{[}4{]} D. Chen, D. Xu, H. Li, N. Sebe, and X. Wang, ``Group Consistent
Similarity Learning via Deep CRF for Person Re-Identification,'' in
\emph{Proc. IEEE/CVF Conference on Computer Vision and Pattern
Recognition}, 2018, pp. 8649--8658, doi: 10.1109/CVPR.2018.00902.

{[}5{]} Y. Wu, O. E. F. Bourahla, X. Li, F. Wu, Q. Tian, and X. Zhou,
``Adaptive Graph Representation Learning for Video Person
Re-Identification,'' \emph{IEEE Trans. Image Process.}, vol. 29, pp.
8821--8830, 2020, doi: 10.1109/TIP.2020.3001693.

{[}6{]} M. Ye, L. Zheng, J. Li, and P. C. Yuen, ``Dynamic Label Graph
Matching for Unsupervised Video Re-Identification,'' in \emph{Proc. ICCV}, 2017, doi:
10.1109/ICCV.2017.550.

{[}7{]} M. Ye, J. Shen, G. Lin, T. Xiang, L. Shao, and S. C. H. Hoi,
``Deep Learning for Person Re-identification: A Survey and Outlook,''
\emph{IEEE Trans. Pattern Anal. Mach. Intell.}, vol. 44, no. 6, 2021, doi:
10.1109/TPAMI.2021.3054775.

{[}8{]} B. Lavi, M. F. Serj, and I. Ullah, ``Survey on Deep Learning
Techniques for Person Re-Identification Task,'' 2018, {[}Online{]}.
Available: https://arxiv.org/abs/1807.05284.

{[}9{]} D. Chicco, ``Siamese Neural Networks: An Overview,''
\emph{Methods Mol. Biol.}, vol. 2190, pp. 73--94, 2021, doi:
10.1007/978-1-0716-0826-5\_3.

{[}10{]} L. Wu, C. Shen, and A. Van Den Hengel, ``PersonNet\,: Person
Re-identification with Deep Convolutional Neural Networks,'' pp. 1--7,
2016, {[}Online{]}. Available: https://arxiv.org/abs/1601.07255.

{[}11{]} H. Luo, Y. Gu, X. Liao, S. Lai, and W. Jiang, ``Bag of Tricks
and A Strong Baseline for Deep Person Re-identification,'' in
\emph{Proc. CVPRW}, 2019, pp. 4320--4329, doi: 10.1109/CVPRW.2019.00190.

{[}12{]} Z. Zhu \emph{et al.}, ``Viewpoint-aware loss with angular
regularization for person re-identification,'' in \emph{Proc. 34th
AAAI Conf. Artif. Intell.}, pp. 13114--13121, 2020, doi:
10.1609/aaai.v34i07.7014.

{[}13{]} A. Schumann and R. Stiefelhagen, ``Person Re-Identification by
Deep Learning Attribute-Complementary Information,'' in \emph{Proc. CVPR}, 2017, 
pp. 20--28, doi: 10.1109/CVPRW.2017.186.

{[}14{]} P. Re-identification, Y. Shen, H. Li, S. Yi, D. Chen, and X.
Wang, ``Person Re-identification with Deep Similarity-Guided Graph
Neural Network,'' in \emph{Proc. ECCV}, 2018, pp. 1--20.

{[}15{]} X. Lan, X. Zhu, and S. Gong, ``Universal Person
Re-Identification,'' 2019, {[}Online{]}. Available:
http://arxiv.org/abs/1907.09511.

{[}16{]} Z. Zeng, Z. Wang, Z. Wang, Y. Zheng, Y. Y. Chuang, and S.
Satoh, ``Illumination-Adaptive Person Re-Identification,'' \emph{IEEE
Trans. Multimed.}, vol. 22, no. 12, pp. 3064--3074, 2020, doi:
10.1109/TMM.2020.2969782.

{[}17{]} F. Xiong, M. Gou, O. Camps, and M. Sznaier, ``Person
re-identification using kernel-based metric learning methods,''
\emph{Lect. Notes Comput. Sci.}, vol. 8695, no. 7, pp.
1--16, 2014, doi: 10.1007/978-3-319-10584-0\_1.

{[}18{]} W. Zheng, S. Gong, and T. Xiang, ``Towards Open-World Person
Re-Identification by One-Shot Group-based Verification,'' \emph{IEEE Trans. Pattern Anal. Mach. Intell.}, vol. 38, no.
3, pp. 591--606, 2016, doi: 10.1109/TPAMI.2015.2453984.

{[}19{]} G. Chen, T. Gu, J. Lu, J. A. Bao, and J. Zhou, ``Person
Re-Identification via Attention Pyramid,'' \emph{IEEE Trans. Image
Process.}, vol. 30, pp. 7663--7676, 2021, doi: 10.1109/TIP.2021.3107211.

{[}20{]} F. M. Khan and F. Bremond, ``Person Re-identification for
Real-world Surveillance Systems,'' 2016, {[}Online{]}. Available:
http://arxiv.org/abs/1607.05975.

{[}21{]} D. Gray and H. Tao, ``Viewpoint Invariant Pedestrian
Recognition with an Ensemble of Localized Features,'' in \emph{Proc. ECCV}, pp.
262--275, 2008.

{[}22{]} T. B. Sebastian, P. H. Tu, J. Rittscher, and R. Hartley,
``Person Reidentification Using Spatiotemporal Appearance,'' in
\emph{Proc. CVPR}, 2006, pp. 1528--1535, doi: 10.1109/CVPR.2006.223.

{[}23{]} L. Nanni, M. Munaro, S. Ghidoni, E. Menegatti, and S. Brahnam,
``Ensemble of different approaches for a reliable person
re-identification system,'' \emph{Appl. Comput. Informatics}, vol. 12,
no. 2, pp. 142--153, 2016, doi:
https://doi.org/10.1016/j.aci.2015.02.002.

{[}24{]} W. S. Zheng, S. Gong, and T. Xiang, ``Person re-identification
by probabilistic relative distance comparison,'' \emph{Proc. CVPR}, pp. 649--656, 2011,
doi: 10.1109/CVPR.2011.5995598.

{[}25{]} M. M. Kalayeh, E. Basaran, M. E. Kamasak, and M. Shah, ``Human
Semantic Parsing for Person Re-identification,'' in \emph{Proc. CVPR},
2018, pp. 1062--1071.

{[}26{]} H. Park and B. Ham, ``Relation Network for Person
Re-identification,'' in \emph{Proc. 34th AAAI Conference on
Artificial Intelligence}, pp. 11839--11847, 2020.

{[}27{]} R. Quan, X. Dong, Y. Wu, L. Zhu, and Y. Yang, ``Auto-ReID:
Searching for a Part-Aware ConvNet for Person Re-Identification,'' in
\emph{Proc. ICCV}, pp. 3750--3759, 2019.

{[}28{]} L. Zheng, L. Shen, L. Tian, S. Wang, J. Wang, and Q. Tian,
``Scalable Person Re-identification\,: A Benchmark,'' \emph{Proc. ICCV}, pp.
1116--1124, 2015.

{[}29{]} Y. Fu \emph{et al.}, ``Horizontal pyramid matching for person
re-identification,'' in \emph{Proc. 33rd AAAI Conf. Artif. Intell.}, pp. 8295--8302, 2019, doi:
10.1609/aaai.v33i01.33018295.

{[}30{]} K. He, G. Gkioxari, P. Dollár, and R. Girshick, ``Mask R-CNN,''
\emph{IEEE Trans. Pattern Anal. Mach. Intell.}, vol. 42, no. 2, pp.
386--397, 2020, doi: 10.1109/TPAMI.2018.2844175.

{[}31{]} K. Gong, X. Liang, D. Zhang, X. Shen, and L. Lin, ``Look into
Person: Self-supervised Structure-sensitive Learning and a new benchmark
for human parsing,'' in \emph{Proc. CVPR}, pp. 6757--6765, 2017, doi:
10.1109/CVPR.2017.715.

{[}32{]} J. Zhao, J. Li, Y. Cheng, T. Sim, S. Yan, and J. Feng,
``Understanding humans in crowded scenes: Deep nested adversarial
learning and a new benchmark for multi-human parsing,'' in \emph{Proc. ACM Multimed. Conf.}, 
pp. 792--800, 2018, doi: 10.1145/3240508.3240509.

{[}33{]} C. Su, J. Li, S. Zhang, J. Xing, W. Gao, and Q. Tian,
``Pose-Driven Deep Convolutional Model for Person Re-identification,'' in
\emph{Proc. ICCV}, pp. 3980--3989, 2017, doi: 10.1109/ICCV.2017.427.

{[}34{]} P. Li, Y. Xu, Y. Wei, and Y. Yang, ``Self-Correction for Human
Parsing,'' \emph{IEEE Trans. Pattern Anal. Mach. Intell.}, vol. 44, no.
6, pp. 3260--3271, 2020, doi: 10.1109/TPAMI.2020.3048039.

{[}35{]} U. Park, A. K. Jain, I. Kitahara, K. Kogure, and N. Hagita,
``ViSE: Visual search engine using multiple networked cameras,'' in
\emph{Proc. ICPR}, pp. 1204--1207, 2006, doi: 10.1109/ICPR.2006.1176.

{[}36{]} G. Wyszecki and W. S. Stiles, \emph{Color Science: Concepts and
Methods, Quantitative Data and Formulae}, 2nd Editio. Wiley, 2000.

{[}37{]} Y. Rubner, C. Tomasi, and L. J. Guibas, ``The Earth Mover's
Distance as a Metric for Image Retrieval,'' in \emph{Proc. IJCV}, pp. 1--20,
2000.

{[}38{]} T. Chavdarova \emph{et al.}, ``WILDTRACK: A Multi-camera HD
Dataset for Dense Unscripted Pedestrian Detection,'' in \emph{Proc. CVPR}, 
pp. 5030--5039, 2018, doi: 10.1109/CVPR.2018.00528.

{[}39{]} S. H. Cha and S. N. Srihari, ``On measuring the distance
between histograms,'' \emph{Pattern Recognit.}, vol. 35, no. 6, pp.
1355--1370, 2002, doi: 10.1016/S0031-3203(01)00118-2.

{[}40{]} M. F. Demirci, A. Shokoufandeh, and S. J. Dickinson, ``Skeletal
shape abstraction from examples,'' \emph{IEEE Trans. Pattern Anal. Mach.
Intell.}, vol. 31, no. 5, pp. 944--952, 2009, doi:
10.1109/TPAMI.2008.267.

{[}41{]} X. Shu and X. J. Wu, ``A novel contour descriptor for 2D shape
matching and its application to image retrieval,'' \emph{Image Vis.
Comput.}, vol. 29, no. 4, pp. 286--294, 2011, doi:
10.1016/j.imavis.2010.11.001.

{[}42{]} S. Thewsuwan and K. Horio, ``Texture-Based Features for
Clothing Classification via Graph-Based Representation,'' \emph{J. signal process.}, vol. 22, no.
6, pp. 299--305, 2018.

{[}43{]} W. Zhou, A. Ahrary, and S. I. Kamata, ``Fase description with
local patterns: An application to face recognition,'' \emph{IEEE Trans. Pattern Anal. Mach.
Intell.}, vol. 28, no. 12, pp. 2037--2041, 2006, doi: 10.1109/TPAMI.2006.244.

{[}44{]} O. Barkan, J. Weill, L. Wolf, and H. Aronowitz, ``Fast high
dimensional vector multiplication face recognition,'' in \emph{Proc. ICCV}, pp. 1960--1967, 2013, doi:
10.1109/ICCV.2013.246.

{[}45{]} B. H. Shekar and B. Pilar, ``Shape representation and
classification through pattern spectrum and local binary pattern - A
decision level fusion approach,'' in \emph{Proc. ICSIP}, pp. 218--224, 2014, doi:
10.1109/ICSIP.2014.41.

{[}46{]} T. X. and X. W. W. Li, R. Zhao, ``DeepReID: Deep Filter Pairing
Neural Network for Person Re-identification,'' in \emph{Proc. CVPR}, pp. 152--159, 2014, doi:
10.1109/CVPR.2014.27.

{[}47{]} Z. Zhong, L. Zheng, D. Cao, and S. Li, ``Re-ranking person
re-identification with k-reciprocal encoding,'' in \emph{Proc. CVPR}, pp.
3652--3661, 2017, doi: 10.1109/CVPR.2017.389.

{[}48{]} R. Quispe and H. Pedrini, ``Top-dB-net: Top dropblock for
activation enhancement in person re-identification,'' in \emph{Proc. ICPR}, pp. 2980--2987, 2020, doi:
10.1109/ICPR48806.2021.9412017.

{[}49{]} H. X. Yu, W. S. Zheng, A. Wu, X. Guo, S. Gong, and J. H. Lai,
``Unsupervised person re-identification by soft multilabel learning,'' in
\emph{Proc. CVPR}, pp. 2143--2152, 2019, doi: 10.1109/CVPR.2019.00225.

{[}50{]} Z. Zheng, L. Zheng, and Y. Yang, ``A discriminatively learned
CNN embedding for person reidentification,'' \emph{ACM Trans. Multimed.
Comput. Commun. Appl.}, vol. 14, no. 1, pp. 1--10, 2017, doi:
10.1145/3159171.

{[}51{]} D. Li, X. Chen, Z. Zhang, and K. Huang, ``Learning deep
context-aware features over body and latent parts for person
re-identification,'' in \emph{Proc. CVPR}, pp. 7398--7407, 2017, doi:
10.1109/CVPR.2017.782.

{[}52{]} M. Wang, B. Lai, J. Huang, X. Gong, and X.-S. Hua,
``Camera-aware Proxies for Unsupervised Person Re-Identification,''
2021, {[}Online{]}. Available: http://arxiv.org/abs/2012.10674.

{[}53{]} R. Quispe and H. Pedrini, ``Improved person re-identification
based on saliency and semantic parsing with deep neural network
models,'' \emph{Image Vis. Comput.}, vol. 92, 2019, doi:
10.1016/j.imavis.2019.07.009.

{[}54{]} F. Zheng \emph{et al.}, ``Pyramidal person re-identification
via multi-loss dynamic training,'' in \emph{Proc. CVPR}, pp. 8506--8514, 2019,
doi: 10.1109/CVPR.2019.00871.

{[}55{]} M. Wieczorek, B. Rychalska, and J. Dabrowski, ``On the
Unreasonable Effectiveness of Centroids in Image Retrieval,'' in
\emph{Proc. ICONIP}, pp. 212--223, 2021, doi: 10.1007/978-3-030-92273-3\_18.

{[}56{]} W. Li, X. Zhu, and S. Gong, ``Harmonious Attention Network for
Person Re-identification,'' in \emph{Proc. CVPR}, pp. 2285--2294, 2018, doi:
10.1109/CVPR.2018.00243.

{[}57{]} Y. Wang \emph{et al.}, ``Resource Aware Person
Re-identification Across Multiple Resolutions,'' in \emph{Proc. CVPR}, pp. 8042--8051,
2018, doi: 10.1109/CVPR.2018.00839.

{[}58{]} K. Zhou, Y. Yang, A. Cavallaro, and T. Xiang, ``Omni-scale
feature learning for person re-identification,'' in \emph{Proc. ICCV}, pp. 3701--3711, 
2019, doi: 10.1109/ICCV.2019.00380.

{[}59{]} C. Ding, K. Wang, P. Wang, and D. Tao, ``Multi-Task Learning
with Coarse Priors for Robust Part-Aware Person Re-Identification,''
\emph{IEEE Trans. Pattern Anal. Mach. Intell.}, vol. 44, no. 3, pp.
1474--1488, 2022, doi: 10.1109/TPAMI.2020.3024900.

{[}60{]} A. Benzine, M. E. A. Seddik, and J. Desmarais, ``Deep Miner: A
Deep and Multi-branch Network which Mines Rich and Diverse Features for
Person Re-identification,'' 2021, {[}Online{]}. Available:
http://arxiv.org/abs/2102.09321.

{[}61{]} F. Herzog, X. Ji, T. Teepe, S. Hörmann, J. Gilg, and G. Rigoll,
``Lightweight Multi-Branch Network for Person Re-Identification,'' in
\emph{Proc. ICIP}, pp. 1129--1133, 2021, doi: 10.1109/ICIP42928.2021.9506733.

{[}62{]} Y. Jiang, S. Yang, H. Qiu, W. Wu, C. C. Loy, and Z. Liu,
``Text2Human: Text-Driven Controllable Human Image Generation,''
\emph{ACM Trans. Graph.}, vol. 41, no. 4, 2022, doi:
10.1145/3528223.3530104.

{[}63{]} Y. Wang, L. Li, J. Yang, and J. Dang, ``Person
re-identification based on attention mechanism and adaptive weighting,''
\emph{Dyna}, vol. 96, no. 2, pp. 186--193, 2021, doi: 10.6036/9981.

\end{document}